\documentclass[a4paper,fleqn]{cas-sc}

\usepackage[numbers]{natbib}
\usepackage{multirow}
\usepackage{pifont}
\usepackage{amsfonts}
\usepackage{comment}
\usepackage{afterpage}

\def\tsc#1{\csdef{#1}{\textsc{\lowercase{#1}}\xspace}}
\tsc{WGM}
\tsc{QE}
\tsc{EP}
\tsc{PMS}
\tsc{BEC}
\tsc{DE}

\begin{document}
\let\WriteBookmarks\relax
\def\floatpagepagefraction{1}
\def\textpagefraction{.001}
\shorttitle{Enhancement for Scribble-Supervised Segmentation of Medical
Images}
\shortauthors{Hoang-An Vo et~al.}

\title [mode = title]{QMaxViT-Unet+: A Query-Based MaxViT-Unet with Edge Enhancement for Scribble-Supervised Segmentation of Medical Images}                      

\tnotemark[1]
\tnotetext[1]{This research was supported by The VNUHCM-University of Information Technology's Scientific Research Support Fund}

\author[1,2]{Thien B. Nguyen-Tat}[orcid=https://orcid.org/0000-0002-4809-7126]
\cormark[1]
\ead{thienntb@uit.edu.vn}
\credit{Conceptualization, methodology, writing - review \& editing, supervision}

\author[1,2]{Hoang-An Vo}
\credit{Software, writing - original draft, visualization}

\author[1,2]{Phuoc-Sang Dang}
\credit{Software, writing - original draft}

\affiliation[1]{organization={University of Information Technology},
                city={Ho Chi Minh City},
                country={Vietnam}}
\affiliation[2]{organization={Vietnam National University},
                city={Ho Chi Minh City},
                country={Vietnam}}

\cortext[cor1]{Corresponding author}

\begin{abstract}
The deployment of advanced deep learning models for medical image segmentation is often constrained by the requirement for extensively annotated datasets. Weakly-supervised learning, which allows less precise labels, has become a promising solution to this challenge. Building on this approach, we propose QMaxViT-Unet+, a novel framework for scribble-supervised medical image segmentation. This framework is built on the U-Net architecture, with the encoder and decoder replaced by Multi-Axis Vision Transformer (MaxViT) blocks. These blocks enhance the model's ability to learn local and global features efficiently. Additionally, our approach integrates a query-based Transformer decoder to refine features and an edge enhancement module to compensate for the limited boundary information in the scribble label. We evaluate the proposed QMaxViT-Unet+ on four public datasets focused on cardiac structures, colorectal polyps, and breast cancer: ACDC, MS-CMRSeg, SUN-SEG, and BUSI. Evaluation metrics include the Dice similarity coefficient (DSC) and the 95th percentile of Hausdorff distance (HD95). Experimental results show that QMaxViT-Unet+ achieves 89.1\% DSC and 1.316mm HD95 on ACDC, 88.4\% DSC and 2.226mm HD95 on MS-CMRSeg, 71.4\% DSC and 4.996mm HD95 on SUN-SEG, and 69.4\% DSC and 50.122mm HD95 on BUSI. These results demonstrate that our method outperforms existing approaches in terms of accuracy, robustness, and efficiency while remaining competitive with fully-supervised learning approaches. This makes it ideal for medical image analysis, where high-quality annotations are often scarce and require significant effort and expense. The code is available at \url{https://github.com/anpc849/QMaxViT-Unet}.


\end{abstract}

\begin{graphicalabstract}
     \begin{figure}[htbp]
       \centering
       \includegraphics[width=\linewidth]{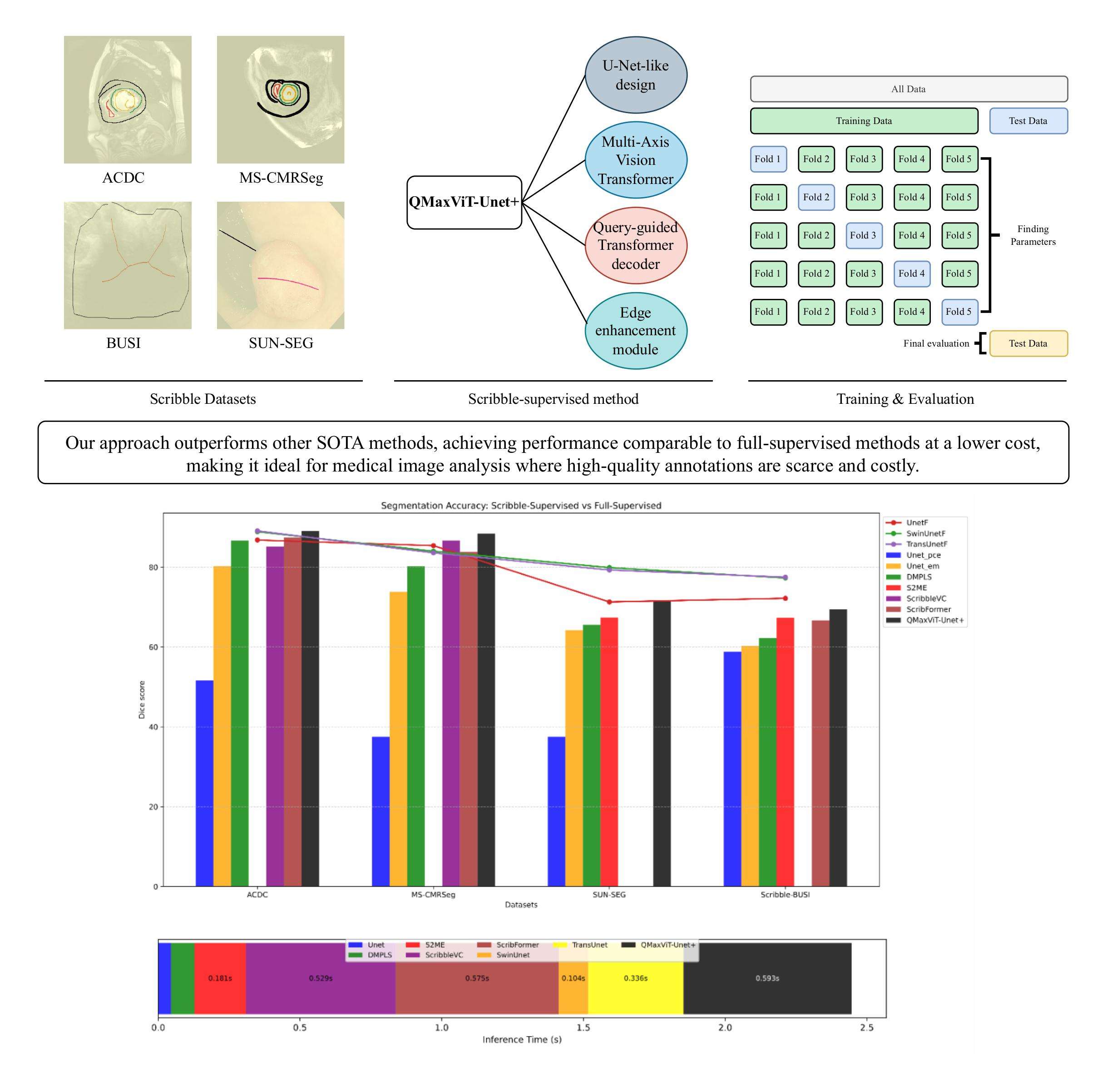}
     \end{figure}
\end{graphicalabstract}

\begin{highlights}
	\item Use U-Net design with MaxViT blocks to learn local and global features efficiently.
    \item Integrate query-based Transformer decoder to refine features and add variants.
    \item Enhance edge details with an Edge module addressing scribble label limitations.
    \item Evaluate the method on four diverse medical imaging datasets.
    \item Discuss performance gains from novel design components in segmentation tasks.
\end{highlights}


\maketitle

\section{Introduction}
Recent advances in deep learning have significantly boosted segmentation techniques by effectively learning complex features directly from medical imaging data. This progress has led to more accurate and detailed segmentation of organ structures, helping improve patient outcomes. Traditional deep learning models, such as the U-Net architecture and its variants, have been foundational in the field of medical image segmentation \cite{ronneberger2015u,oktay2018attention,zhou2018unet++,isensee2018nnu}. While convolutional layers in U-Net excel at capturing image details, they show weaknesses in accessing global and long-range semantic information. Consequently, numerous studies have combined convolutional layers with transformer layers to address these shortcomings \cite{li2022rt,jiang2022transcunet, NGUYENTAT2024100528}. With the development of the Vision Transformers (ViTs) \cite{dosovitskiy2020image}, transformers have been increasingly applied to medical image segmentation \cite{chen2021transunet,zhou2021nnformer,hatamizadeh2021swin,hatamizadeh2022unetr,peiris2022robust,xu2023levit}. As deep learning approaches gain popularity, transfer learning plays a crucial role in improving the performance of medical segmentation models by leveraging pre-trained networks. Techniques such as STU-Net \cite{huang2023stu} have shown the ability of large-scale supervised pre-training and transfer learning in medical contexts. These approaches achieved robust performance, especially when trained on large datasets with dense labeling. However, generating such datasets is resource-intensive, requiring significant time and expertise in medical annotation to produce the dense labels necessary for training models.

To address these challenges, the field has increasingly adopted weakly-supervised learning strategies. These strategies use less accuracy, and more readily obtainable labels, such as points, bounding boxes, and scribbles. The application of scribble-supervised learning has emerged as a promising method to reduce the training costs associated with densely annotated datasets while maintaining high performance. This technique has been effectively applied in various vision tasks, including object detection and semantic segmentation. \cite{ji2019scribble} proposed a scribble-based hierarchical weakly-supervised learning model for brain tumor segmentation. This model combines two types of weak labels for training: scribbles indicating the whole tumor and healthy brain tissue, along with global labels denoting the presence of each substructure. \cite {luo2022scribble} employed a dual-branch network with one encoder and two slightly different decoders for segmentation and dynamically mixed the two decoders’s predictions to generate pseudo labels for supervision. According to the author, this design offers two advantages. First, it mitigates the inherent weaknesses of pseudo-labels in a single-branch network, as the outputs from the two branches differ due to feature perturbation. Second, it allows for pseudo-label generation through the ensemble of two outputs without requiring the training of two separate networks, thereby improving the encoder's ability to extract features through dual supervision. To address the challenge of learning global features in convolutional layers, \cite {li2023scribblevc} proposed the ScribbleVC which utilized a traditional CNN-Transformer hybrid architecture to simultaneously learn local and global features. Additionally, ScribbleVC incorporated class information extracted from scribble labels to create a multimodal information enhancement module. The combination of CNN and Transformer architectures allowed the model to effectively learn both local and global features. Furthermore, \cite {li2024scribformer} continued this work by incorporating an attention-guided class activation map (ACAM) branch into the CNN-Transformer framework to further improve performance.

However, scribble annotations often lack the detailed boundary information that dense labels provide. To address this limitation, \cite {zhang2020weakly} proposed an auxiliary edge detection task to explicitly localize object boundaries and a gated structure-aware loss to constrain the recovery of structural details within a defined scope. Additionally, \cite {zhang2022cyclemix} introduced a framework called CycleMix, which adopted a mixup strategy combined with a dedicated design for random occlusion. This approach enabled incremental and decremental adjustments of scribbles. CycleMix improved the training objective with consistency losses, penalizing inconsistent segmentations and thereby improving boundary information, which resulted in a significant improvement in segmentation performance. Furthermore, \cite {wang2023s} designed a framework known as Spatial-Spectral Dual-Branch Mutual Teaching and Entropy-Guided Pseudo Label Ensemble Learning (S2ME). This framework takes advantage of the natural compatibility of features extracted from both spatial and spectral domains, encouraging cross-space consistency through collaborative optimization. It also introduced a novel adaptive pixel-wise fusion technique to refine the boundaries based on entropy guidance derived from the spatial and spectral branches.

Based on previous research and our observations, we outline the key challenges in scribble-supervised segmentation. Firstly, the application of transfer learning in these contexts remains under-explored, offering an opportunity for substantial improvements in reducing training costs and enhancing model performance. Secondly, while traditional CNN-Transformer hybrids can successfully learn both local and global features at the same time, they often perform worse than traditional convolutional networks if not pre-trained. According to \cite{tu2022maxvit}, this gap comes from the strong capacity of Transformers, which have less inductive bias and are likely to overfit when trained on smaller datasets. Additionally, acquiring global interactions through full attention at early or high-resolution stages in a hierarchical network is computationally intensive due to the quadratic complexity of the attention mechanism. Lastly, scribble annotations lack detailed boundary information compared to dense annotations, which can lead to poor segmentation performance. This is particularly critical in the medical field, where accurate boundary segmentation is essential.

In this study, we propose a novel framework for scribble-supervised medical image segmentation named QMaxViT-Unet+ to address these issues. In particular, QMaxViT-Unet+ utilizes MaxViT blocks, as proposed by \cite{tu2022maxvit}, to replace the encoder and decoder blocks in the U-Net architecture. By employing these pre-trained MaxViT blocks as the backbone, we can effectively handle the first two issues in scribble-supervised segmentation. Additionally, our approach integrates a query-based Transformer decoder to refine features and build an auxiliary decoder, which addresses the issue mentioned above in a single-branch network and enhances the output variations. The last issue is addressed by integrating an Edge enhancement module, which incorporates edge information extracted from the MaxViT encoder into the MaxViT decoder blocks. This methodology not only reduces the dependence on extensive annotated data but also aims to closely match the segmentation quality of fully-supervised models, thereby making high-quality medical image segmentation more accessible and less labor-intensive. Our contributions can be summarized as follows:
\begin{itemize}
 	\item To the best of our knowledge, our approach is the first to use transfer learning with a pre-trained Vision Transformer for scribble-supervised medical image segmentation, rather than training from scratch. It leverages a pre-trained MaxViT as the backbone and builds upon a U-Net type design to capture both detailed high-resolution spatial information and global context from images through an attention mechanism with linear complexity, in contrast to the quadratic complexity of traditional attention mechanisms.
	\item We introduce a query-based Transformer decoder and an Edge enhancement module to enhance the performance of segmentation results. The query-based Transformer decoder refines encoder features and serves as an auxiliary decoder to address the inherent weaknesses of pseudo-labels in a single-branch network and adds output variants. The Edge enhancement module is specifically designed to enhance the boundary information that is often missing in scribble labels compared to dense labels.
 	\item Various experiments have evaluated the proposed QMaxViT-Unet+ on four benchmark datasets: ACDC, MS-CMRSeg, SUN-SEG, and BUSI. The results demonstrate that our method outperforms state-of-the-art methods, achieving DSC of 89.1\%, 88.4\%, 71.4\%, and 69.4\% and HD95 of 1.316mm, 2.226mm, 4.996mm and 50.122mm, respectively. The results also highlight the significant contributions of the query-based Transformer decoder, and the Edge enhancement module to the overall performance of QMaxViT-Unet+.
\end{itemize}

The structure of our work is organized as follows: Section~\hyperref[sec:materials_and_methods]{\ref*{sec:materials_and_methods}} provides a detailed description of the datasets employed in this study, as well as the methodology and training strategies utilized. Section~\hyperref[sec:experiments_and_results]{\ref*{sec:experiments_and_results}} presents the experimental settings and a thorough analysis of the results obtained. Furthermore, we conduct a series of ablation studies in Section~\hyperref[sec:discussion]{\ref*{sec:discussion}} to rigorously evaluate the contribution of individual components of our proposed model, discuss its limitations, and propose directions for future research. Finally, we summarize the paper in Section~\hyperref[sec:conclusion]{\ref*{sec:conclusion}}.

\section{Materials and Methods}\label{sec:materials_and_methods}
\subsection{Datasets}
\begin{figure}[htbp]
  \centering
  \includegraphics[width=\linewidth]{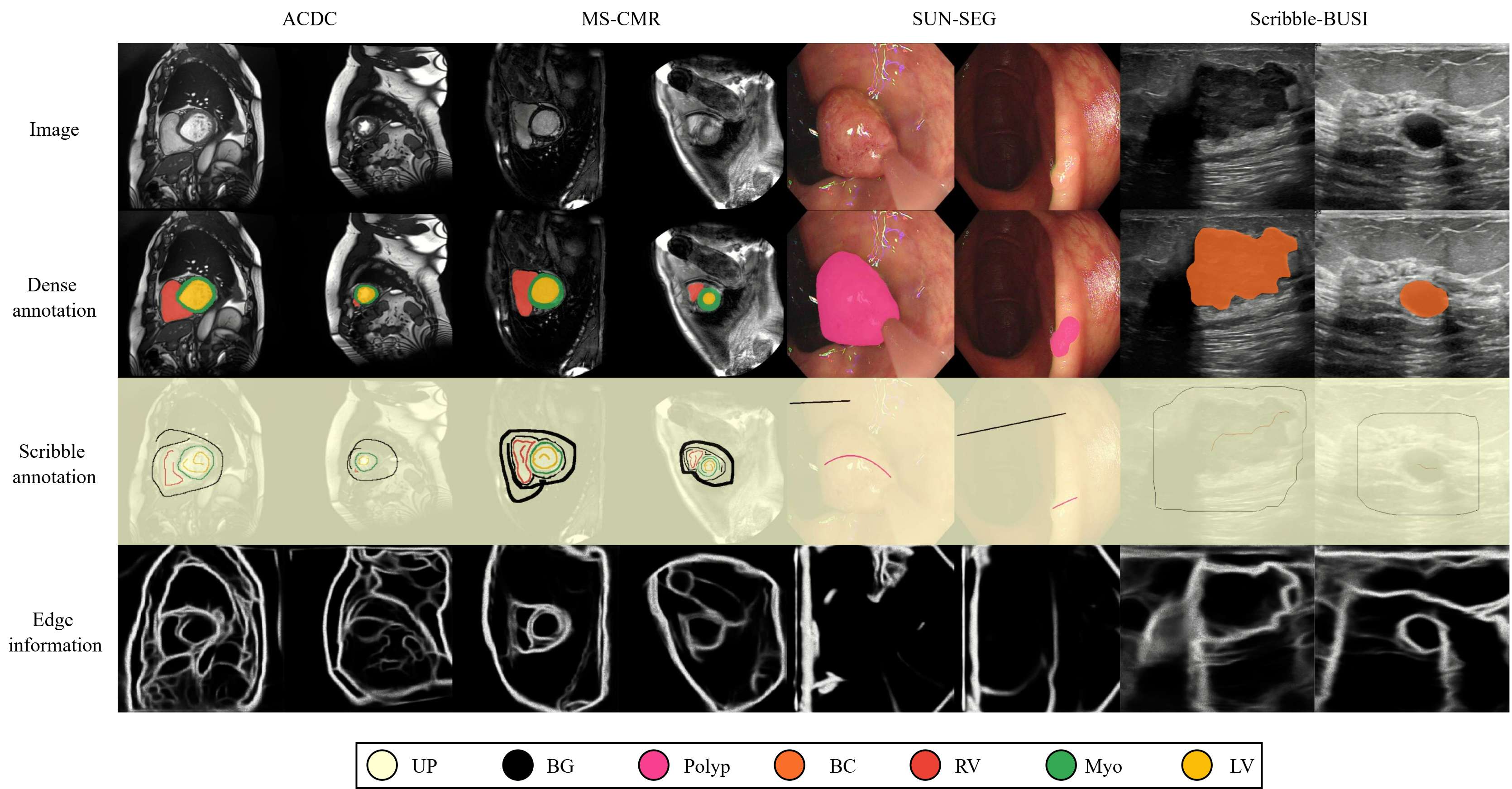}
  \caption{Examples of dense, scribble annotations and edge information from the ACDC, MS-CMRSeg, SUN-SEG, and BUSI dataset. UP, BG, Polyp, BC, RV, Myo, and LV represent the
  unannotated, background, colon polyp, breast cancer, right ventricle, myocardium, and left ventricle pixels, respectively.}
  \label{fig:data_visual}
\end{figure}

We evaluate our proposed method on four medical datasets: ACDC, MS-CMRSeg, SUN-SEG, and BUSI. Visual representations and statistical summaries of these datasets are shown in Fig. \ref{fig:data_visual} and Tab. \ref{dataset:metadata}, respectively.

The \textbf{ACDC} dataset consists of cine-MRI scans from 150 patients, with manual scribble annotations for the left ventricle (LV), right ventricle (RV), and myocardium (MYO) provided by experts as detailed in \cite{acdc_ref1,acdc_ref2}. Following \cite{luo2022scribble}, we employ a five-fold cross-validation strategy, utilizing the first 100 patients, while reserving the remaining 50 for testing in each fold. The \textbf{MS-CMRSeg} dataset \cite{mscmr_ref1,mscmr_ref2} consists of late gadolinium enhancement (LGE) MRI scans from 45 cardiomyopathy patients, with scribble annotations for LV, MYO, and RV provided by \cite{zhang2022cyclemix}. Each patient has a volumetric scan consisting of approximately 15 slices (images), resulting in a total of around 686 images. Following the approach in \cite{zhang2022cyclemix,zhang2022shapepu}, we randomly split the dataset at the patient level by selecting 25 patients (382 images) for training, 5 patients (75 images) for validation, and 15 patients (229 images) for testing. Because only the training subset has the required scribble annotations, we use these 382 training images for model learning, while the remaining subsets are reserved for validation and final performance evaluation. The \textbf{SUN-SEG} dataset with scribble annotations \cite{ji2022video,ji2021progressively,fan2020pranet} is a subset of SUN database \cite{misawa2021development}. This dataset contains 100 different polyp video cases. To reduce data redundancy and memory consumption \cite{wang2023s}, we only select the first frame of every five consecutive frames in each case. The dataset is then split into 70, 10, and 20 cases for training, validation, and testing, respectively. The \textbf{BUSI} dataset \cite{al2020dataset}, which includes breast ultrasound images collected in 2018 from women aged 25 to 75 years old. Normal samples without segmentation masks were excluded. Since no manual scribbles are available, scribble annotations were generated using WSL4MIS code\footnote{\url{https://github.com/HiLab-git/WSL4MIS}} provided by HiLab at UESTC, which extracts the two largest connected components from binary masks and refines branching structures. Similar to ACDC, we utilize five-fold cross-validation to evaluate the performance of this dataset. For edge information extraction, we use the pre-trained model \cite{zhou2023treasure} to perform inference on the images and obtain edge masks (treated as ground truth), respectively.

\begin{table}[width=.9\linewidth,cols=8,pos=htbp]
  \caption{The coverage of scribble annotations (calculated based on dense annotations) by regions and the number of training samples vary across the datasets. Although the MS-CMRSeg dataset has the fewest training samples, it shows the highest scribble coverage. In contrast, the SUN-SEG dataset, despite having a larger number of training samples, exhibits lower scribble coverage.}\label{dataset:metadata}
  \begin{tabular*}{\tblwidth}{@{}LLLLLLLLL@{}}
  \toprule
  \multirow{2.5}{*}{Dataset} & \multicolumn{6}{c}{Scribble coverage (\%)} & \multirow{2.5}{*}{Modality} & \multirow{2.5}{*}{Training samples} \\
  \cmidrule(lr){2-7}
    & RV &  MYO & LV & Polyp & BC & BG &  & \\
  \midrule
  ACDC & 17.58 & 20.16 & 10.48 & - & - & 0.75 & MRI & 1522\\
  MS-CMRSeg & \textbf{30.92} & \textbf{38.14} & \textbf{27.67} & - & - & \textbf{5.01} & MRI & 382\\
  SUN-SEG & - &  - & - & 6.29 & - & 0.60 & Endoscopy & \textbf{6677}\\
  BUSI & - & - & - & - & 0.91 & 0.48 & Ultrasound & 413 \\
  \bottomrule
  \end{tabular*}
\end{table}

\subsection{QMaxViT-Unet+ Architecture}

\begin{figure}[htbp]
  \centering
  \includegraphics[width=0.5\linewidth]{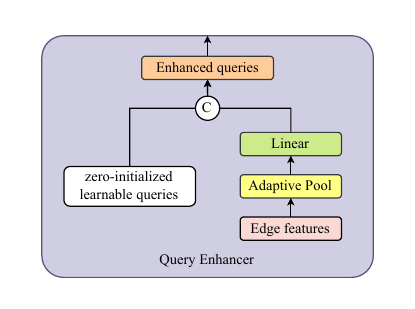}
  \caption{Simple Query enhancer. The edge features extracted from the Edge enhancement module are processed through an adaptive pooling layer and a linear layer. These features are then combined with zero-initialized queries to create improved query representations.}
  \label{fig:query_enhancer}
\end{figure}

\begin{figure}[htbp]
  \centering
  \includegraphics[width=\linewidth]{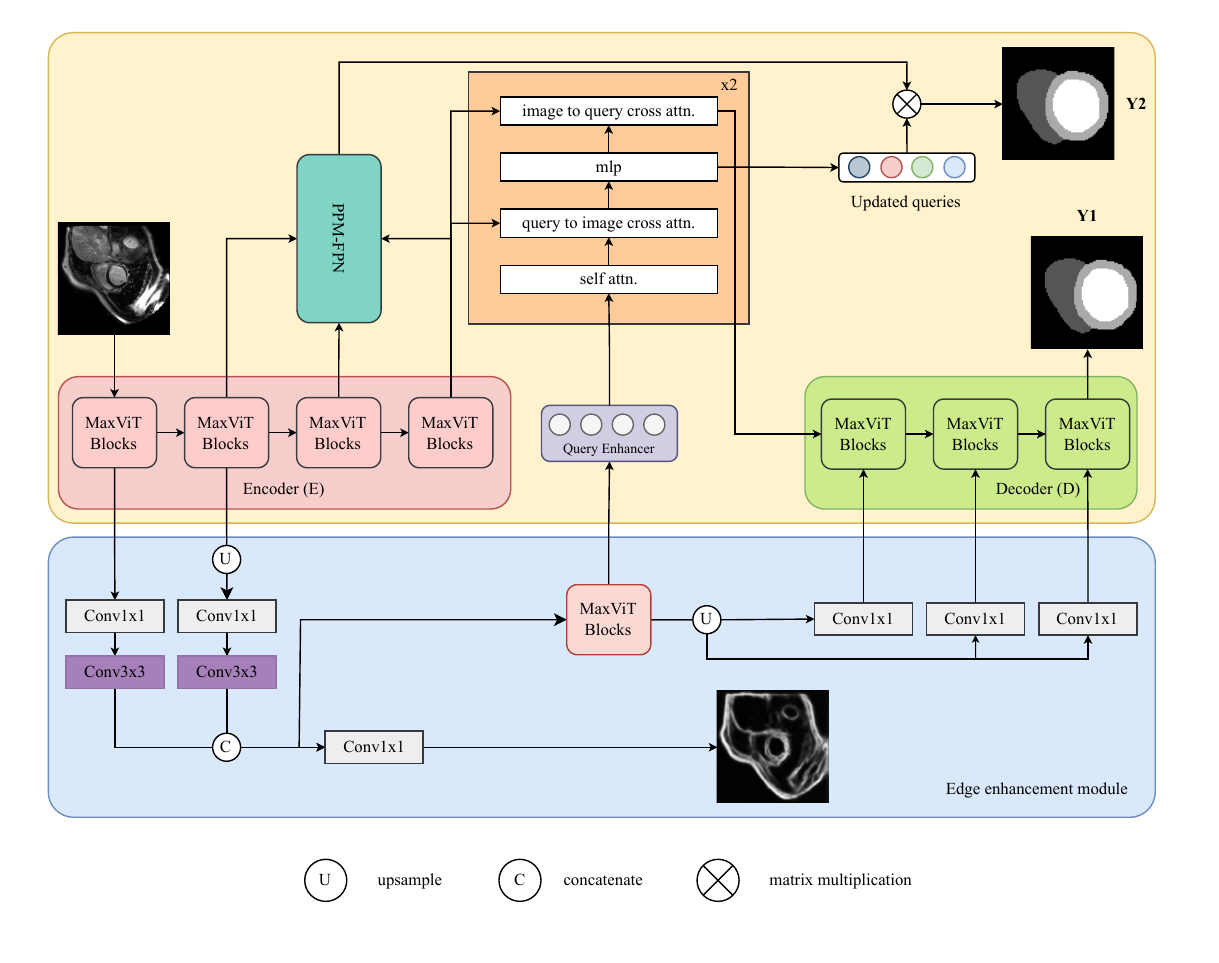}
  \caption{\textbf{QMaxViT-Unet+ architecture.} The proposed architecture is based on the U-Net framework, with the conventional U-Net blocks replaced by MaxViT blocks. To improve segmentation accuracy, we incorporate the PPM-FPN module, the Query-guided Transformer decoder, the Edge enhancement module, and the Query enhancer. For readability, skip connections and positional embeddings are omitted from the diagram.}
  \label{fig:architecture}
\end{figure}

\paragraph{Overview of QMaxViT-Unet+.} The architecture of the QMaxViT-Unet+ framework is presented in Fig. \ref{fig:architecture}. The framework comprises three key components: MaxViT blocks, Edge enhancement module, and query-guided Transformer decoder. Given an input triplet \((x, s, e)\) representing the image, scribble, and ground-truth edge from dataset \(D = \{(x, s, e)_n\}_{n=1}^N\), the QMaxViT-Unet+ first processes the input through four E-blocks to extract both local and global features. Unlike traditional U-Net architectures, our approach routes the last E-block (bottleneck) to the query-guided Transformer decoder to refine attention features before passing them to the first D-block. The last D-block, followed by a conv2d layer, produces the primary segmentation mask \(\mathsf{y}_1\). To further enhance edge information, the features from the first and second E-blocks are fed into the Edge Enhancement module. The enhanced outputs from this module are then combined with the features from the D-blocks, improving the overall segmentation quality. The PPM-FPN module processes the features from the last three E-blocks, improving them before performing a matrix multiplication with the updated queries to produce an auxiliary segmentation mask \(\mathsf{y}_2\). Furthermore, the Query enhancer (Fig. \ref{fig:query_enhancer}) is incorporated to improve the learnable queries, motivated by the findings of \cite{cheng2022masked, guo2024mask2former}, which demonstrate the limitations of zero-initialization. In comparison to existing state-of-the-art CNN-Transformer hybrid networks \cite{li2023scribblevc,li2024scribformer}, QMaxViT-Unet+ not only learns local and global features more efficiently but also surpasses them in segmentation and boundary quality by incorporating edge information and leveraging the query-guided Transformer decoder for refined features representation.

\paragraph{MaxViT-Unet.} Traditional Convolutional Neural Networks (CNNs) excel at capturing local image features but struggle to model long-range dependencies, which are crucial for effective medical image analysis. While Vision Transformers (ViTs) can capture both local and global features, they often suffer from high computational demands and a tendency to overfit, particularly when trained from scratch. The MaxViT block, proposed by \cite{tu2022maxvit} addresses these issues by integrating Mobile Convolution Blocks (MBConv) with computes features and utilizes Squeeze-and-Excitation (SE) attention, Block Attention, and Grid Attention mechanisms, thereby efficiently modeling spatial interactions while maintaining linear complexity. Our proposed QMaxViT-Unet+ framework builds upon the MaxViT-Unet architecture, introduced by \cite{khan2023maxvit}, which employs a U-Net-like design to maintain contextual information via skip connections. The MaxViT-Unet replaces traditional encoder-decoder blocks with MaxViT Stages, each comprising a variable number of MaxViT blocks - a hyperparameter in our model. By combining the U-Net design with MaxViT blocks, our model's ability to handle complex features is greatly improved. Moreover, using ImageNet-1k \cite{deng2009imagenet} pre-training for the backbone allows for transfer learning, which further boosts segmentation accuracy. The backbone processes images of size $3 \times 256 \times 256$. For gray image $1 \times 256 \times 256$, a conv2d layer, followed by BatchNorm \cite{ioffe2015batch} and ReLU activation \cite{agarap2018deep}, is used to project the single grayscale channel to three channels. The backbone generates four feature representations of size $1 \times (96 \times 2^i) \times \frac{H}{2^{2+i}} \times \frac{W}{2^{2+i}}$ across four different stages, where $i = \{0, 1, 2, 3\}$.

\paragraph{Edge Enhancement Module.} A notable limitation of scribble-supervised learning is the lack of boundary information compared to full-supervised learning. Drawing inspiration from \cite{zhang2019net}, we design a straightforward Edge enhancement module that incorporates enhanced edge information into the D-blocks. Our observations suggest that the first two E-blocks typically capture low-level features such as edges. Therefore, we apply this module on top of these early E-blocks. In our Edge enhancement module, the outputs from the second E-block are upsampled to align with the resolution of the outputs from the first E-block. The upsampled features are then fed into a 1x1-3x3 convolutional layer before being concatenated and subsequently divided into two branches. The first branch is passed through a 1x1 convolutional layer to generate edge masks, which are supervised by automatically generated ground truth edges. The second branch is fed into a MaxViT Stage to produce attention maps, which are subsequently used to combine with zero-initialized queries in the Query enhancer (Fig. \ref{fig:query_enhancer}). The attention maps are also upsampled followed by 1x1 convolutional layers to augment the edge information for D-blocks.

\paragraph{Query-guided Transformer decoder.} Inspired by the work of \cite{cheng2021per}, MaskFormer demonstrated the potential of query-guided Transformer decoder in computer vision tasks. This approach leverages powerful relation and attribution modeling to achieve impressive results. \cite{yang2024query} further explored this concept, proposing QuCCeS, a method that effectively applies query-based Transformer for cross-center generalization in medical image segmentation, especially in situations with limited data. Inspired by these advancements, we incorporate a query-guided Transformer decoder into our model to enhance segmentation quality. We define a set of queries $q = {q_1, q_2, ...q_i}$, where $i$ represents the number of categories in the segmentation task. Unlike QuCCeS, which employed zero-initialization for the queries, we use the simple Query enhancer (Fig. \ref{fig:query_enhancer}) that combines zero-initialized queries with the attention maps generated by the Edge enhancement module to produce enhanced learnable queries. Specifically, we concatenate the zero-initialized queries with a shape of $(\text{num\_class}, \text{D}_{\text{E4}}/2)$ with the output of the Query enhancer, which has a shape of $(\text{num\_class}, \text{D}_{\text{E4}}/2)$, to generate the final learnable queries with a final shape of $(\text{num\_class}, \text{D}_{\text{E4}})$, where $\text{D}_{\text{E4}}$ denotes the hidden dimension of the output features from the last E-block. The enhanced learnable queries and features output from the last E-block, are then fed into the Transformer decoder \cite{kirillov2023segment} to refine and enhance their attention to important features, resulting in the updated queries. Additionally, we observe that the highest level features of the last E-block are insufficient for generating a segmentation mask, so we employ a variant of PPM-FPN \cite{cheng2022sparse} instead of the vanilla FPN \cite{lin2017feature}, which incorporates a pyramid pooling module \cite{zhao2017pyramid} to enlarge the receptive field and fuse multi-scale features. Finally, we perform matrix multiplication to assign attention weights from the updated learnable queries to multi-scale features, thereby generating the auxiliary segmentation mask $\mathsf{y}_2$.

\subsection{Loss Functions}

\begin{figure}[htbp]
  \centering
  \includegraphics[width=\linewidth]{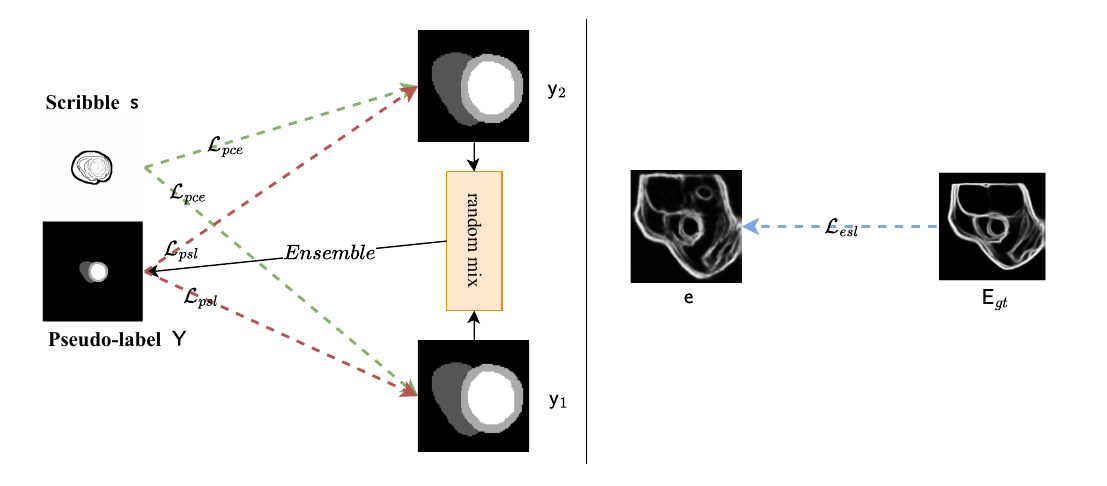}
  \caption{Loss Functions}
  \label{fig:loss_supervision}
\end{figure}

\paragraph{Scribble-supervised Loss.} We apply the partial cross-entropy loss for scribble-supervised learning, which ignores unlabeled pixels (UP) in the scribble annotation. The loss for scribble supervision ($\mathcal{L}_{ssl}$) with a sample $(x, s, e)$ is formulated as:
\begin{equation}
  \mathcal{L}_{ssl}(s, \mathsf{y}_1, \mathsf{y}_2) = \frac{1}{2} \left[ \mathcal{L}_{pce}(\mathsf{y}_1, s) + \mathcal{L}_{pce}(\mathsf{y}_2, s) \right] ; \quad
  \mathcal{L}_{pce}(\mathsf{y}, s) = - \sum_{i \in S \setminus \text{UP}} y_i \log(p_i)
\end{equation}
where \( S \) is the set of labels used for calculating cross-entropy, excluding the label UP (unknown pixel).

\paragraph{Pseudo-supervised Loss.} Following \cite{luo2022scribble}, we generate the hard pseudo labels by mixing two predictions dynamically $\mathsf{y}_1$ and $\mathsf{y}_2$. The pseudo-supervised loss ($\mathcal{L}_{psl}$) is formulated as:
\begin{equation}
  \mathcal{L}_{psl}(s, \mathsf{y}_1, \mathsf{y}_2) = \frac{1}{2} \left[ \mathcal{L}_{dice}(\mathsf{y}_1, \mathsf{Y}) + \mathcal{L}_{dice}(\mathsf{y}_2, \mathsf{Y}) \right]; \quad
  \mathsf{Y} = \texttt{argmax}(\alpha \cdot \mathsf{y}_1 + \beta \cdot \mathsf{y}_2), \quad \alpha, \beta \in (0,1)
\end{equation}
where $\mathcal{L}_{dice}$ is the Dice function, $\mathsf{Y}$ is the pseudo label, and $\alpha$ is a dynamically generated random number in the range $(0, 1)$ in each iteration, while $\beta$ is set to $1 - \alpha$.

\paragraph{Edge-supervised Loss.} The Edge enhancement module is supervised using the Mean Squared Error (MSE) regression loss function, which measures the discrepancy between the predicted edge values and the ground truth edge values. The edge-supervised loss function, denoted as $\mathcal{L}_{esl}$, is formally defined as:
\begin{equation}
	\mathcal{L}_{esl} = \frac{1}{n} \sum_{p=1}^{n} \left(\text{EEM}(p) - E_{gt}(p)\right)^2
\label{eq:boundaryloss_function}
\end{equation}
where $\text{EEM}(p)$ represents the predicted edge value output by the Edge enhancement module, and $E_{gt}(p)$ is the corresponding ground truth edge value at pixel $p$, $n$ is the total number of pixels in the image.

\paragraph{Final Loss.} The final loss function to train our QMaxViT-Unet+ is formulated as:
\begin{equation}
    \mathcal{L}_{total}=\lambda_1 \times \mathcal{L}_{ssl}+\lambda_2 \times \mathcal{L}_{psl}+\lambda_3 \times \mathcal{L}_{esl}
\label{eq:totalloss_function}
\end{equation}
where $\lambda_{1-3}$ are the weights of each part of the loss to balance different supervised losses. Fig. \ref{fig:loss_supervision} illustrates the supervised loss functions employed in our method.

\section{Experiments and Results}\label{sec:experiments_and_results}
\subsection{Implementation Details}
The model was implemented using PyTorch\footnote{\url{https://pytorch.org/}} and trained on a single NVIDIA RTX 4090 GPU. We performed on-the-fly data augmentation by randomly rotating and flipping the images during training. The augmented images were resized to $256 \times 256$ pixels before being fed into the network. The learning rate was set to 1e-3, with a weight decay of 0.01. The model was trained using the AdamW optimizer for 200 epochs, with a CosineAnnealingLR scheduler\footnote{\url{https://pytorch.org/docs/stable/generated/torch.optim.lr_scheduler.CosineAnnealingLR.html}} applied during the experiments. The weights ($\lambda_1, \lambda_2, \lambda_3$) in Equation \ref{eq:totalloss_function} were empirically set to (1, 0.5, 0.2). During inference, only the resized $256 \times 256$ images were required. For all datasets, the Dice Score (DSC) and the 95th percentile of Hausdorff Distance (HD95) were used as evaluation metrics.
\subsection{Results}

To fully evaluate the segmentation performance of our proposed QMaxViT-Unet+ model, we conducted a comparative analysis with state-of-the-art (SOTA) methods on four benchmark datasets: ACDC, MS-CMRSeg, SUN-SEG and BUSI. The results are presented in Tab. \ref{tab:results_acdc}, \ref{tab:results_mscmr}, \ref{tab:results_sunseg} and \ref{tab:results_busi}, respectively. Each table is divided into two sections: scribble-supervised methods and full-supervised methods. We did not include comparisons with data augmentation strategies such as Puzzle Mix \cite{kim2020puzzle}, Cutout \cite{devries2017improved}, MixUp \cite{zhang2017mixup}, and CycleMix \cite{zhang2022cyclemix} in the scribble-supervised setting, as they exhibited slower training convergence and inferior performance compared to current SOTA methods.

On the ACDC dataset, QMaxViT-Unet+ significantly outperformed all previous models. It surpassed ScribbleVC by 3.9\% DSC and 0.462 mm HD95, and ScribFormer by 1.7\% DSC and 5.15 mm HD95. Notably, our model achieved a 2.2\% higher DSC than ScribFormer in the myocardium region. Furthermore, when compared to full-supervised methods, our model achieved comparable performance to the top-performing TransUnet, but with lower annotation costs. Although our results were slightly lower than TransUnet's for the myocardium (in DSC) and left ventricle (in HD95), the results remained competitive. In the case of the smaller MS-CMRSeg dataset, where many SOTA methods experienced a performance drop sometimes even under-performing traditional CNN models like Unet, our QMaxViT-Unet+ consistently outperformed them, demonstrating robust performance with minimal degradation compared to training on the ACDC dataset. Specifically, it exceeded ScribbleVC by 1.9\% DSC and 7.22mm HD95, and ScribFormer by 4.6\% DSC and 3.832mm HD95. Compared to fully-supervised methods, QMaxViT-Unet+ outperformed all, particularly surpassing $Unet_F$ by 3\% DSC and 1.064mm HD95. On the largest dataset, SUN-SEG, QMaxViT-Unet+ surpassed the current SOTA method, S2ME, by 4\% DSC, although it showed a slightly higher HD95. When compared to fully-supervised methods, our model achieved results comparable to $Unet_F$ at a lower annotation effort, though it still fell short of the performance of SwinUnet and TransUnet. On the BUSI dataset, QMaxViT-Unet+ demonstrated superior performance compared to prior methods. It outperformed S2ME by 2.1\% DSC and 19.004 mm HD95, and ScribFormer by 2.7\% DSC and 12.545 mm HD95. Compared to fully-supervised methods, QMaxViT-Unet+ achieved competitive results, with a Dice score approaching $Unet_F$ at a significantly lower annotation cost. Although it slightly underperformed compared to SwinUnet and TransUnet in HD95, its performance remained commendable. Overall, these findings underscore the significant potential of the proposed scribble-supervised model in medical image segmentation, achieving comparable performance to fully-supervised methods while minimizing annotation cost.

\begin{table}[width=.9\linewidth,cols=9,pos=htbp]
  \centering
  \caption{Performance comparison of QMaxViT-Unet+ with other SOTA methods on the ACDC dataset, using Dice Score and HD95 metrics. Results are averaged from 5-fold cross-validation.}\label{tab:results_acdc}
  \begin{tabular*}{\tblwidth}{@{\extracolsep{\fill}} lcccc cccc@{}}
    \toprule
    \multirow{2}{*}{Methods} & \multicolumn{4}{c}{$\text{Dice}\!\uparrow$} & \multicolumn{4}{c}{$\text{HD95}\!\downarrow$} \\
    \cmidrule(lr){2-5} \cmidrule(lr){6-9}
     & RV & MYO & LV & Avg & RV & MYO & LV & Avg \\
    \midrule
    \textbf{Scribbles} & & & & & & & & \\
    Unet$_{pce}$ \cite{lin2016scribblesup}& .476 & .473 & .600 & .516 & 118.306 & 107.953 & 112.435 & 112.898 \\
    Unet$_{em}$ \cite{grandvalet2004semi}& .801 & .759 & .848 & .803 & 57.948 & 58.129 & 53.815 & 56.631 \\
    DMPLS \cite{luo2022scribble}& .868 & .829 & .904 & .867 & 4.503 & 10.049 & 13.785 & 9.446 \\
    ScribbleVC \cite{li2023scribblevc}& .830 & .839 & .886 & .852 & 2.340 & 1.258 & 1.735 & 1.778 \\
    ScribFormer \cite{li2024scribformer}& .869 & .842 & .911 & .874 & 4.305 & 7.816 & 7.276 & 6.466 \\
    QMaxViT-Unet+ (\textbf{ours}) & \textbf{.884} & .864 & \textbf{.924} & \textbf{.891} & 1.663 & \textbf{1.125} & \textbf{1.159} & \textbf{1.316} \\
    \textbf{Masks} & & & & & & & & \\
    Unet$_F$ \cite{ronneberger2015u}& .857 & .836 & .912 & .868 & 1.913 & 1.506 & 2.439 & 1.953 \\
    SwinUnet \cite{hatamizadeh2021swin}& .879 & .868 & .921 & .889 & 3.478 & 2.292 & 4.434 & 3.401 \\
    TransUnet \cite{chen2021transunet}& .883 & \textbf{.870} & .920 & .891 & \textbf{1.613} & 1.183 & 1.907 & 1.568 \\
    \bottomrule
  \end{tabular*}
\end{table}

\begin{table}[width=.9\linewidth,cols=9,pos=htbp]
  \centering
  \caption{Performance comparison of QMaxViT-Unet+ with other SOTA methods on the MS-CMRSeg dataset, using Dice Score and HD95 metrics.}\label{tab:results_mscmr}
  \begin{tabular*}{\tblwidth}{@{\extracolsep{\fill}} lcccc cccc@{}}
  \toprule
  \multirow{2}{*}{Methods} & \multicolumn{4}{c}{$\text{Dice}\!\uparrow$} & \multicolumn{4}{c}{$\text{HD95}\!\downarrow$} \\
  \cmidrule(lr){2-5} \cmidrule(lr){6-9}
   & RV & MYO & LV & Avg & RV & MYO & LV & Avg\\
  \midrule
  \textbf{Scribbles} &  &  &  &  &  &  &  & \\
  Unet$_{pce}$ \cite{lin2016scribblesup} & .298 & .401 & .426 & .375 & 251.070 & 212.160 & 250.217 & 237.816 \\
  Unet$_{em}$ \cite{grandvalet2004semi} & .786 & .665 & .762 & .738 & 140.178 & 170.602 & 176.114 & 162.298 \\
  DMPLS \cite{luo2022scribble} & .810 & .756 & .839 & .802 & 100.271 & 59.467 & 92.378 & 84.039 \\
  ScribbleVC \cite{li2023scribblevc} & .868 & .829 & .904 & .867 & 4.503 & 10.049 & 13.785 & 9.446 \\
  ScribFormer \cite{li2024scribformer} & .827 & .789 & .897 & .838 & 11.936 & 2.804 & 3.435 & 6.058 \\
  QMaxViT-Unet+ (\textbf{ours}) & \textbf{.888} & \textbf{.842} & \textbf{.923} & \textbf{.884} & \textbf{2.079} & \textbf{2.252} & \textbf{2.346} & \textbf{2.226}\\
  \textbf{Masks} &  &  &  &  &  &  &  & \\
  Unet$_F$ \cite{ronneberger2015u} & .848 & .805 & .909 & .854 & 4.031 & 2.924 & 2.915 & 3.290 \\
  SwinUnet \cite{hatamizadeh2021swin}& .837 & .790 & .892 & .840 & 3.385 & 18.849 & 14.535 & 12.256 \\
  TransUnet \cite{chen2021transunet}& .818 & .794 & .895 & .836 & 3.749 & 3.460  & 6.959 & 4.723 \\
  \bottomrule
  \end{tabular*}
\end{table}

\begin{table}[width=.9\linewidth,cols=3,pos=htbp]
  \centering
  \caption{Performance comparison of QMaxViT-Unet+ with other SOTA methods on the SUN-SEG dataset, using Dice Score and HD95 metrics.}\label{tab:results_sunseg}
  \begin{tabular*}{\tblwidth}{@{} LCC@{} }
  \toprule
  Method & $\text{Dice}\!\uparrow$ & $\text{HD95}\!\downarrow$ \\
  \midrule
  \textbf{Scribbles} & & \\
  Unet$_{pce}$ \cite{lin2016scribblesup}& .375 & 237.816 \\
  Unet$_{em}$ \cite{grandvalet2004semi} & .642 & 5.277 \\
  USTM \cite{liu2022weakly}& .654 & 5.207 \\
  CPS \cite{chen2021semi}& .658 & 5.092 \\
  DMPLS \cite{luo2022scribble}& .656 & 5.208 \\
  S2ME \cite{wang2023s}& .674 & 4.583 \\
  QMaxViT-Unet+ (\textbf{ours}) & .714 & 4.996 \\
  \textbf{Masks} & & \\
  Unet$_F$ \cite{ronneberger2015u}& .713 & 4.405 \\
  SwinUnet \cite{hatamizadeh2021swin}& \textbf{.799} & 3.817 \\
  TransUnet \cite{chen2021transunet}& .793 & \textbf{3.716} \\
  \bottomrule
  \end{tabular*}
\end{table}

\begin{table}[width=.9\linewidth,cols=3,pos=htbp]
  \centering
  \caption{Performance comparison of QMaxViT-Unet+ with other SOTA methods on the BUSI dataset, using Dice Score and HD95 metrics.}\label{tab:results_busi}
  \begin{tabular*}{\tblwidth}{@{} LCC@{} }
  \toprule
  Method & $\text{Dice}\!\uparrow$ & $\text{HD95}\!\downarrow$ \\
  \midrule
  \textbf{Scribbles} & & \\
  Unet$_{pce}$ \cite{lin2016scribblesup}& .588 & 102.983 \\
  Unet$_{em}$ \cite{grandvalet2004semi} & .603 & 89.007 \\
  DMPLS \cite{luo2022scribble}& .623 & 78.442 \\
  ScribFormer \cite{li2024scribformer} & .667& 62.667\\
  S2ME \cite{wang2023s}& .673 & 69.126 \\
  QMaxViT-Unet+ (\textbf{ours}) & .694 & 50.122 \\
  \textbf{Masks} & & \\
  Unet$_F$ \cite{ronneberger2015u}& .722 & 68.512 \\
  SwinUnet \cite{hatamizadeh2021swin}& .773 & \textbf{40.868} \\
  TransUnet \cite{chen2021transunet}& \textbf{.775} & 40.907 \\
  \bottomrule
  \end{tabular*}
\end{table}

\afterpage{
    \begin{figure}[htbp]
      \centering
      \includegraphics[width=\linewidth]    {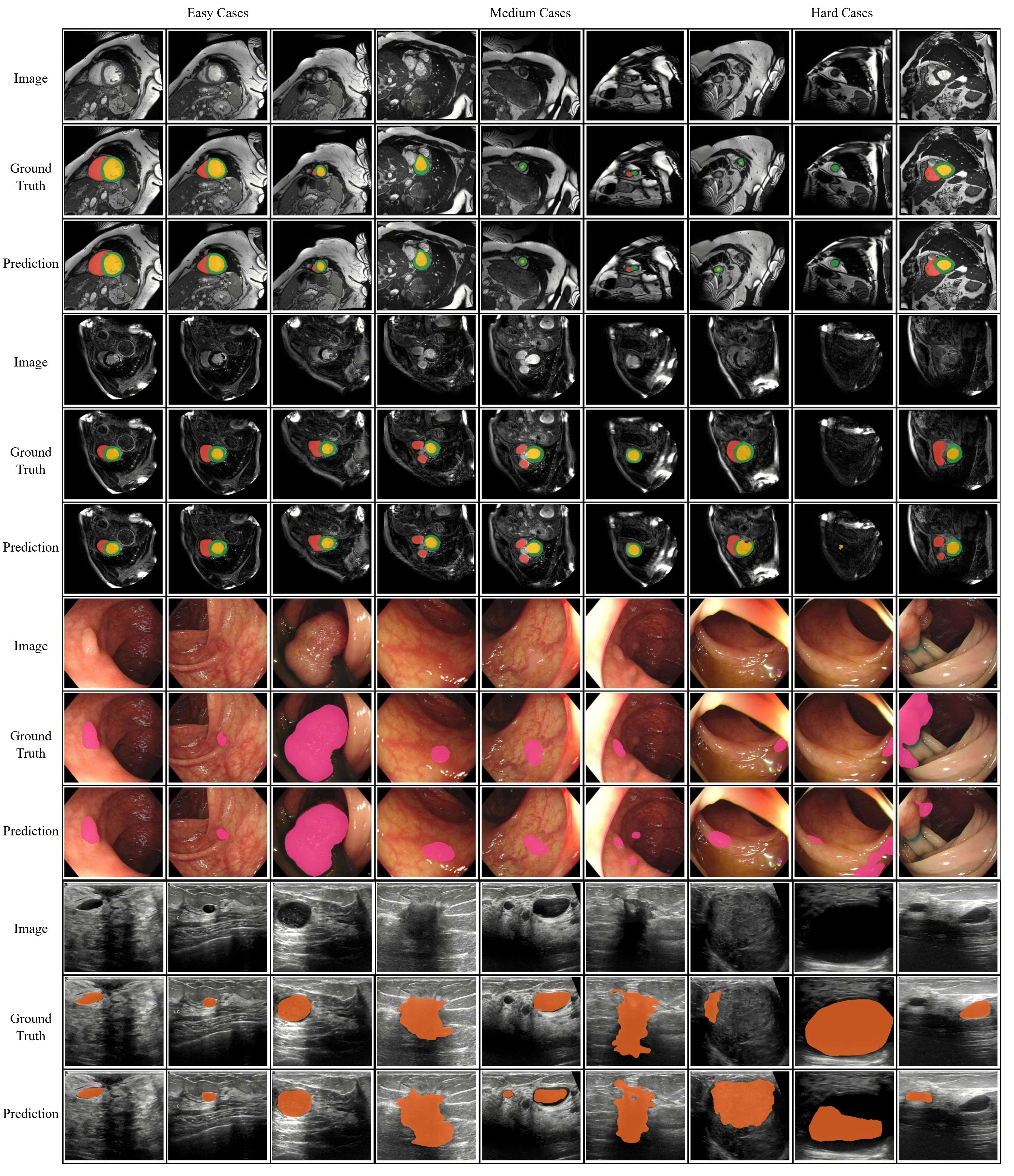}
      \caption{Segmentation results across ACDC, MS-CMRSeg, SUN-SEG and BUSI datasets on cases of varying difficulty.}
      \label{dis:vizsegresults}
    \end{figure}
    \clearpage
    }
    
\section{Discussion}\label{sec:discussion}
This section aims to demonstrate the effectiveness of the QMaxViT-Unet+ design on scribble-supervised medical image segmentation tasks. We conducted a series of diagnostic experiments using the ACDC and MS-CMRSeg datasets. These experiments were designed to evaluate various aspects of our method, including model visualization \ref{dis:vizsegresults}, effectiveness of loss functions \ref{dis:loss_functions}, contribution of different model components \ref{dis:model_components}, cross-dataset performance analysis \ref{dis:cross_performance}, and model complexity \ref{dis:model_complexity}. These results provide a comprehensive understanding of QMaxViT-Unet+'s capabilities and its robustness.

\paragraph{Visualization and Analysis of Segmentation Performance.} Fig. \ref{dis:vizsegresults} illustrates the segmentation results of the QMaxViT-Unet+ model across four datasets, categorized into easy, medium, and hard cases. The model performed with high accuracy in easy cases, with predictions closely aligning with the ground truth in all datasets. For medium cases, the model showed strong performance, particularly in detecting small regions in the cardiac datasets, although some false positives occurred, especially in the SUN-SEG and BUSI dataset, likely due to visual similarities between different regions. But as the cases became more complex, the model's performance declined, struggling with the demarcation of intricate structures in cardiac regions, large polyps, and breast cancer areas, due to the complex structure and variability of these features. Overall, while the model performs well in easy and most medium cases, challenges remain in handling complex structures, image quality, and scribble annotation quality, which impact boundary information and overall performance. This underscores the need for further model refinement to improve its effectiveness in more complex scenarios.

\paragraph{Effectiveness of Loss Functions.} This experiment demonstrates the importance of balancing the weights of different loss components in the QMaxViT-Unet+ model. As shown in Tab. \ref{dis:loss_functions}, we observed that assigning equal weights to all loss components (Set \#1) resulted in fairly good segmentation performance. However, adjusting the weights to prioritize $\lambda_1$ and $\lambda_3$ enhanced segmentation quality, as evidenced by the improved average DSC, though it also led to an increase in HD95 (Sets \#2 and \#3). This happened because increasing the weight on $\lambda_3$ can introduce noise into the model, potentially due to the imperfect edge masks generated by the pre-trained model \cite{zhou2023treasure}, which led to a higher average HD95. These edge masks, although treated as ground truth, may contain errors or inconsistencies that can mislead the model during training. By reducing the weight on $\lambda_3$ (Set \#4), we effectively mitigated this issue and achieved the highest average DSC and the most accurate boundary predictions, as indicated by the lowest average HD95.

\begin{table}[width=.9\linewidth,cols=6,pos=htbp]
  \centering
  \caption{Impact of varying weight settings ($\lambda_1$, $\lambda_2$, $\lambda_3$) on the performance of QMaxViT-Unet+ with the MS-CMRSeg dataset, using Dice Score and HD95 metrics.}\label{dis:loss_functions}
  \begin{tabular*}{\tblwidth}{@{} LLCCCC@{} }
  \toprule
  Set & $\lambda_1$ & $\lambda_2$ & $\lambda_3$ & $\text{Avg}_{\text{dice}}\!\!\uparrow$ &  $\text{Avg}_{\text{hd95}}\!\!\downarrow$ \\
  \midrule
  \#1 & 0.5 & 0.5 & 0.5 & .854 & 2.988\\
  \#2 & 0.7 & 0.2 & 0.4 & .877 & 8.838\\
  \#3 & 0.8 & 0.4 & 0.3 & .880 & 3.059\\
  \#4 &1 & 0.5 & 0.2 & \textbf{.884}  & \textbf{2.226} \\
  \bottomrule
  \end{tabular*}
\end{table}

\paragraph{Effectiveness of Model Components.}
The ablation study presented in Table \ref{dis:model_components} provides valuable insights into the effectiveness of individual components in our proposed QMaxViT-Unet+ model. Note that models integrated with query component but lacking edge component use zero-initialized learnable queries and models without $\mathsf{y}_2$ exclude the PPM-FPN module pathway. The table is divided into two sections: the first section(first four rows) evaluates our model with the single-decoder (main decoder), while the second section (last four rows) assesses the performance with the dual-decoder. In the single-decoder settings, the baseline model without query and edge components achieved an average DSC of 75.2\%. The addition of edge and query components led to a notable improvement in average Dice and HD95 scores. Notably, incorporating both edge and query components simultaneously greatly boosted performance, with an average DSC of 81.8\% and an average HD95 score of 139.566mm, even with a single decoder. In the two-decoder settings, the baseline model without query and edge components achieved an average DSC of 87.3\% and an average HD95 score of 6.749mm, demonstrating the effectiveness of our model with two decoders as explored in \cite{luo2022scribble}. Adding the edge component gave a marginally higher average DSC, accompanied by a significant reduction in the average HD95 score to 2.508mm, underscoring the importance of edge information in enhancing boundary segmentation. Although the addition of the query component resulted in a higher DSC compared to the edge component, it led to a worse HD95 score. This observation suggests that while the query component effectively refines features, thereby enhancing segmentation accuracy, the edge component remains essential for precise boundary segmentation. In the end, our QMaxViT-Unet+ model, equipped with the dual-decoder and both query and edge components, achieved the highest average DSC of 88.4\% and the lowest average HD95 score of 2.226mm. Additionally, in both single-decoder and dual-decoder setups, models with both parts (query + edge) did the best. This might have been due to how these components worked together, possibly aided by the Query enhancer. However, we don't specifically look at the Query enhancer's effect in this study, so future experiments will need to explore this more.

\begin{figure}[htbp]
  \centering
  \includegraphics[width=\linewidth]{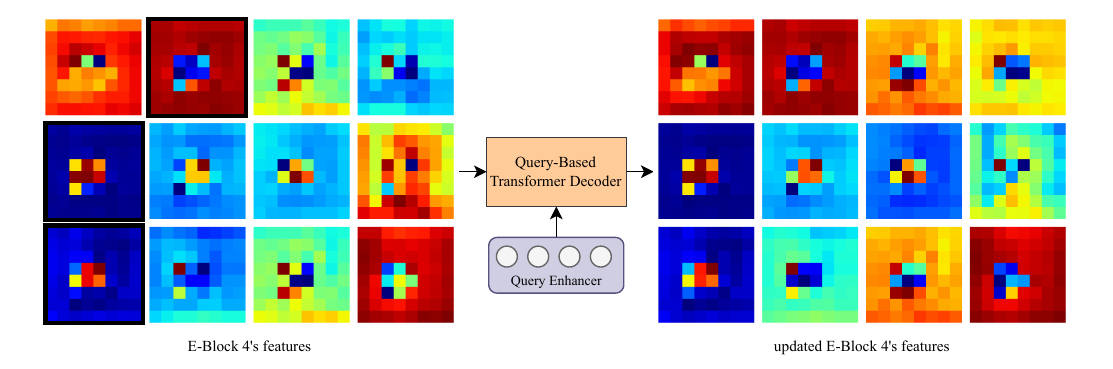}
  \caption{Features visualization of the last E-block before and after refinement by the query-based Transformer decoder. Note that the full feature set consists of 768 features, but only a subset is shown here for readability.}
  \label{dis:e4_block_features}
\end{figure}

\begin{figure}[htbp]
  \centering
  \includegraphics[width=\linewidth]{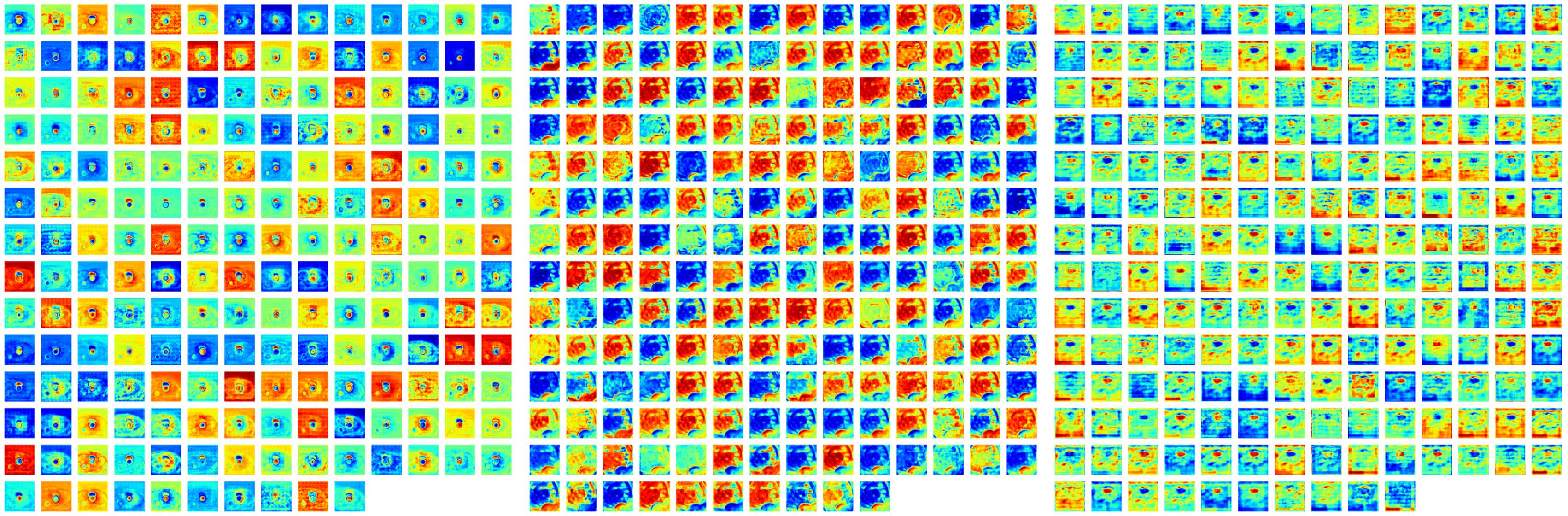}
  \caption{Visualization of attention maps extracted from the MaxViT Stage of the Edge enhancement module, which highlights the model's focus on relevant features in Cardiac datasets (left), Polyp dataset (middle) and BUSI dataset (right).}
  \label{dis:attention_map}
\end{figure}

We further explored how our method worked by visualizing the feature representations extracted from the QMaxViT-Unet+ model during inference. As illustrated in Fig. \ref{dis:e4_block_features}, the features map with a black border around showed that features looked similar before and after refinement, but the previously unclear features became more focused and better defined. This refinement demonstrates that the query-based Transformer decoder effectively focuses on the most relevant features and filters out unnecessary information. Additionally, the attention maps shown in Fig. \ref{dis:attention_map} demonstrate that our model effectively highlights the most important features in medical images. However, the model faced greater difficulty in capturing attention in the more complex SUN-SEG and BUSI dataset compared to the ACDC and MS-CMRSeg datasets.

\begin{table}[width=.9\linewidth,cols=5,pos=htbp]
	\centering
	\caption{Impact of the dual-decoder ($\mathsf{y}_2$), query-based Transformer (query) and Edge enhancement module (edge) on QMaxViT-Unet+ Performance with the MS-CMRSeg dataset, using Dice Score and HD95 metrics.}\label{dis:model_components}
	\begin{tabular*}{\tblwidth}{@{} LCCCC@{} }
		\toprule
		$\mathsf{y}_2$ & query & edge & $\text{Avg}_{\text{dice}}\!\!\uparrow$ &  $\text{Avg}_{\text{hd95}}\!\!\downarrow$ \\
		\midrule
		\ding{55} & \ding{55} & \ding{55} & .752 & 178.712 \\ 
		\ding{55} & \ding{55} & \ding{51} & .772 & 173.583 \\ 
		\ding{55} & \ding{51} & \ding{55} & .783 & 167.621 \\ 
		\ding{55} & \ding{51} & \ding{51} & .818 & 139.566 \\ 
		\ding{51} & \ding{55} & \ding{55} & .873 & 6.749 \\ 
		\ding{51} & \ding{55} & \ding{51} & .878 & 2.508 \\ 
		\ding{51} & \ding{51} & \ding{55} & .879 & 2.611 \\ 
		\ding{51} & \ding{51} & \ding{51} & \textbf{.884} & \textbf{2.226} \\ 
		\bottomrule
	\end{tabular*}
\end{table}

\newpage
\paragraph{Model Generalization.}
We evaluated the robustness of our method through the cross-dataset experiment on ACDC (Cine MRI) and MS-CMRSeg (LGE MRI) datasets, with the results shown in Table \ref{dis:cross_performance}. While SOTA methods excelled when trained separately on each dataset, their performance significantly degraded when cross-evaluated on the test sets. Specifically, the traditional CNN-Transformer hybrid models, ScribbleVC and ScribFormer, exhibited substantial performance drops, with decreases of 31.7\% (85.2\% $\rightarrow$ 53.5\%) and 61.2\% (87.4\% $\rightarrow$ 26.2\%) on ACDC dataset, and 42.2\% (86.7\% $\rightarrow$ 44.5\%) and 25.8\% (83.8\% $\rightarrow$ 58.0\%) on MS-CMRSeg dataset, respectively. Notably, these models performed even worse than CNN architectures, such as Unet$_{em}$ or DMPLS. In contrast, our QMaxViT-Unet+, built on MaxViT Blocks, demonstrated superior performance and generalization capabilities, with smaller drops (16.6\% on ACDC and 18.1\% on MS-CMRSeg). Additionally, our QMaxViT-Unet+ achieved the lowest HD95 score, highlighting the effectiveness of edge information enhancement. This suggests that our model is more suitable for real-world scenarios where MRI images may come from diverse sources.

\begin{table}[width=.9\linewidth,cols=5,pos=htbp]
  \centering
  \caption{Cross-dataset experiment on ACDC and MS-CMRSeg datasets: Performance comparison of segmentation models trained on one dataset and evaluated on the test set of another, which highlights the impact of dataset shift, using Dice Score and HD95 metrics.}\label{dis:cross_performance}
  \begin{tabular*}{\tblwidth}{@{\extracolsep{\fill}} lcc cc@{}}
    \toprule
    \multirow{2}{*}{Methods} & \multicolumn{2}{c}{Train\textsubscript{ACDC} $\rightarrow$ Eval\textsubscript{MS-CMRSeg}} & \multicolumn{2}{c}{Train\textsubscript{MS-CMRSeg} $\rightarrow$ Eval\textsubscript{ACDC}} \\
    \cmidrule(lr){2-3} \cmidrule(lr){4-5}
     & $\text{Avg}_{\text{dice}}\!\!\uparrow$ & $\text{Avg}_{\text{hd95}}\!\!\downarrow$ & $\text{Avg}_{\text{dice}}\!\!\uparrow$ & $\text{Avg}_{\text{hd95}}\!\!\downarrow$ \\
    \midrule
    Unet$_{pce}$ & .433 & 200.346 & .267 & 125.287 \\
    Unet$_{em}$ & .586 & 150.022 & .419 & 111.434 \\
    DMPLS & .655 & 59.933 & .494 & 82.877 \\
    ScribFormer & .580 & 41.250 & .262 & 76.609 \\
    ScribbleVC & .445 & 19.655 & .535 & 28.689 \\
    QMaxViT-Unet+ (\textbf{ours}) & \textbf{.703} & \textbf{10.840} & \textbf{.725} & \textbf{22.217} \\
    \bottomrule
  \end{tabular*}
\end{table}

\paragraph{Model Complexity Comparison.}
To investigate the trade-off between model complexity and performance, we analyzed various models using the ACDC dataset, as shown in Tab. \ref{dis:model_complexity}. We compared parameter counts, Multiply-Accumulate operations (MACs), and average inference times across 100 trials. While CycleMix, which uses a CNN backbone, exhibited the lowest parameter count, its mix strategies added additional complexity. Additionally, traditional CNN-Transformer (ScribbleVC and ScribFormer) architectures demonstrated higher parameter counts and MACs. In contrast, our QMaxViT-Unet+ model, with 109.01 million parameters, is the largest among the evaluated models but has the lowest computational complexity with 39.10G MACs due to its efficient MaxViT Blocks. However, QMaxViT-Unet+'s inference time was slightly higher than others, this might have been due to the model's substantial size. To make the architecture better suited for more scenarios, we are actively refining its design for further optimization.

\begin{table}[width=.9\linewidth,cols=5,pos=htbp]
  \centering
  \caption{Model complexity comparison between our method (QMaxViT-Unet+) and other SOTA methods on the ACDC dataset. The experiment was conducted on a free version virtual machine on Lightning Studios\protect\footnote{\url{https://lightning.ai/studios/}}. Methods marked with an asterisk (*) are fully supervised. Note that UNet is used in both scribble-supervised and full-supervised methods.}\label{dis:model_complexity}
  \begin{tabular*}{\tblwidth}{@{} LLCCC@{} }
  \toprule
  Method & Backbone & Params(M) &  MACs(G) & Inference time (s)\\
  \midrule
  UNet* \cite{ronneberger2015u,lin2016scribblesup,grandvalet2004semi,liu2022weakly} & CNN & 1.81 & 2.97 & 0.045 \\
  DMPLS \cite{luo2022scribble} & CNN & 2.45 & 4.85 & 0.083 \\
  S2ME \cite{wang2023s}& CNN & 8.74 & 15.92 & 0.181 \\
  SwinUnet* \cite{hatamizadeh2021swin} & CNN-Trans & 27.17 & 5.92 & 0.104 \\
  CycleMix \cite{zhang2022cyclemix} & CNN & 25.76 & \textbf{58.76} & 0.562 \\
  CPS \cite{chen2021semi} & CNN & 80.94 & 19.13 & 0.360 \\
  ScribbleVC \cite{li2023scribblevc}& CNN-Trans & 50.27 & 54.48 & 0.529 \\
  ScribFormer \cite{li2024scribformer} & CNN-Trans & 50.43 & 54.74 & 0.575 \\
  TransUnet* \cite{chen2021transunet}& CNN-Trans & 105.28 & 25.37 & 0.336 \\
  QMaxViT-Unet+ \textbf{(ours)}& MaxViT & \textbf{109.01} & 39.10 & \textbf{0.593} \\
  \bottomrule
  \end{tabular*}
\end{table}
\footnotetext{\url{https://lightning.ai/studios/}}

\paragraph{Limitations \& Future works.}
While our proposed model demonstrated promising results, several limitations require further exploration. One significant limitation is the use of imperfect edge masks generated by a pre-trained model, which can introduce noise and potentially mislead the model's attention to irrelevant regions, thereby impacting its performance. Future research directions could involve exploring automated techniques to enhance boundary information while minimizing noise and improving model precision. Additionally, while our model performed well in detecting small regions in medical images, it encountered difficulties in certain cases, as shown in Fig. \ref{dis:vizsegresults}. Addressing this issue may involve experimenting with advanced medical image processing techniques and an in-depth analysis of the relationship between scribble coverage and dataset size. Our observations suggest that the SUN-SEG and BUSI datasets have the two lowest scribble coverages, which may be a contributing factor to the model's poor performance. Another limitation is the lack of the possibility of using cross-validation on the small dataset MS-CMRSeg dataset, which reduces the opportunity for a comprehensive analysis of the model's specific performance characteristics on this dataset and partially limits a thorough assessment of the generalization capability of our proposed method. We hope future research will provide more scribble annotations for MS-CMRSeg or propose new scribble-based datasets to advance model validation and benchmarking in this field. Furthermore, although the model’s inference time is reasonable, its large size requires optimization for wider application. Finally, investigating the potential of unsupervised pre-training of large vision models for medical image segmentation, similar to advancements in large language models, represents a promising direction for future research. 



\section{Conclusion}\label{sec:conclusion}
This paper proposes a novel approach to address the existing limitations in scribble-supervised medical image segmentation tasks, leveraging the efficient Vision Transformer MaxViT blocks. By building upon these blocks, our proposed architecture demonstrates superior efficiency compared to traditional CNN-Transformer hybrid architectures. Additionally, the integration of the query-based Transformer decoder enhances the attention mechanism of the model, while the Edge enhancement module mitigates the lack of boundary information inherent in scribble labels. As a result, our approach outperforms other SOTA methods and achieves comparable performance to fully supervised learning at a lower cost, making it ideal for medical image analysis where high-quality annotations are often scarce and labor-intensive to obtain. This approach shows promise for enhancing computer-aided diagnosis and treatment planning, potentially improving healthcare outcomes through more efficient and accurate image interpretation.

\section*{Data and code availability}\label{secA1}
The corresponding author can provide the datasets and codes generated and/or analyzed during the current work upon reasonable request.

\section*{Ethics approval}\label{secA1}
It is not applicable. The study does not include any medical tests, treatments, or interventions.

\section*{Conflicts of interest}\label{secA1}
The authors have no conflicts of interest to declare that are relevant to the content of this article.

\section*{Acknowledgements} This research was supported by The VNUHCM-University of Information Technology's Scientific Research Support Fund.



\printcredits

\bibliographystyle{unsrt}
\bibliography{cas-refs}






\end{document}